\documentclass[10pt,twocolumn,letterpaper]{article}
\usepackage{wacv}
\usepackage{times}
\usepackage{epsfig}
\usepackage{graphicx}
\usepackage{amsmath}
\usepackage{amssymb}
\usepackage{booktabs}
\usepackage{multirow}
\usepackage{authblk}

\wacvfinalcopy 

\ifwacvfinal
\usepackage[breaklinks=true,bookmarks=false,colorlinks]{hyperref}
\else
\usepackage[pagebackref=true,breaklinks=true,colorlinks,bookmarks=false]{hyperref}
\fi

\begin{document}

\title{MixVPR: Feature Mixing for Visual Place Recognition}

\author[]{Amar Ali-bey}
\author[]{Brahim Chaib-draa}
\author[]{Philippe Gigu\`ere}

\affil[]{Universit\'e Laval, Qu\'ebec, Canada}

\maketitle
\thispagestyle{empty}

\begin{abstract}
Visual Place Recognition (VPR) is a crucial part of mobile robotics and autonomous driving as well as other computer vision tasks. It refers to the process of identifying a place depicted in a query image using only computer vision. At large scale, repetitive structures, weather and illumination changes pose a real challenge, as appearances can drastically change over time. Along with tackling these challenges, an efficient VPR technique must also be practical in real-world scenarios where latency matters.
To address this, we introduce MixVPR, a new holistic feature aggregation technique that takes feature maps from pre-trained backbones as a set of global features. Then, it incorporates a global relationship between elements in each feature map in a cascade of feature mixing, eliminating the need for local or pyramidal aggregation as done in NetVLAD or TransVPR. We demonstrate the effectiveness of our technique through extensive experiments on multiple large-scale benchmarks. Our method outperforms all existing techniques by a large margin while having less than half the number of parameters compared to CosPlace and NetVLAD. We achieve a new all-time high recall@1 score of $94.6\%$ on \mbox{Pitts250k-test}, $88.0\%$ on MapillarySLS, and more importantly, $58.4\%$ on Nordland. 
Finally, our method outperforms two-stage retrieval techniques such as \mbox{Patch-NetVLAD}, TransVPR and SuperGLUE all while being orders of magnitude faster. Our code and trained models are available at \mbox{\url{https://github.com/amaralibey/MixVPR}.}

\end{abstract}

\section{Introduction}
\label{sec:intro}
\begin{figure}[t]
\begin{center}
   \includegraphics[width=1\linewidth]{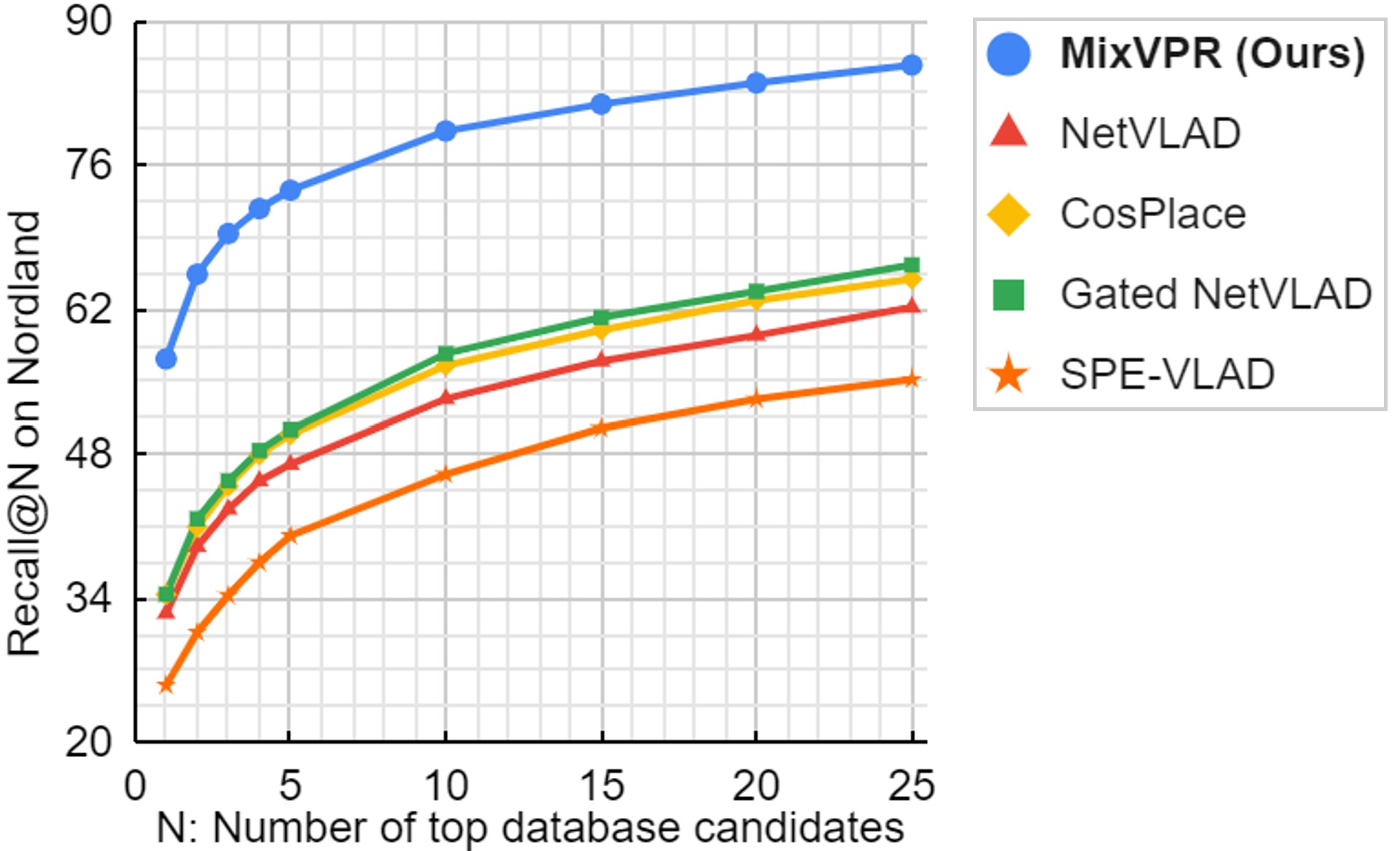}
\end{center}
   \caption{Comparison of performance on the challenging Nordland benchmark. All methods have been trained on the exact same dataset, using the same backbone architecture.}
\label{fig:onecol}
\end{figure}
Visual place recognition (VPR) is an essential part of many robotics \cite{chen2017only, chen2017deep, chen2018learning, garg2019semantic, hausler2019multi, khaliq2019holistic} and computer vision tasks~\cite{arandjelovic2016netvlad, kim2017learned, liu2019stochastic, ge2020self, hausler2021patch, wang2022transvpr, berton2022rethinking} such as autonomous driving~\cite{doan2019scalable}, SLAM~\cite{yadav2022fusion}, image geo-localization~\cite{sarlin2020superglue, cao2020unifying}, virtual reality~\cite{masone2021survey} and 3D~reconstruction~\cite{liu2019lpd}. A visual place recognition system retrieves the location of a given query image by first gathering its visual information into a compact descriptor (image representation), then matching it against a database of references with known geolocations. This task can be extremely challenging due to short term appearance changes (e.g., illumination, occlusion and weather) as well as long term variations (e.g., seasonal changes, construction and vegetation). Therefore, a robust VPR technique should be capable of producing descriptors that are invariant to these changes.

Traditionally, VPR technique used hand-crafted local features such as SIFT \cite{lowe2004distinctive} and SURF~\cite{bay2006surf} which can be further aggregated into a global descriptor that represents the entire image such as Fisher Vectors~\cite{jegou2010aggregating, perronnin2010large}, Bag of Words~\cite{philbin2007object, torii2013visual, galvez2012bags} and Vector of Locally Aggregated Descriptor (VLAD)~\cite{jegou2011aggregating,arandjelovic2013all}. Following the growth of deep learning, where convolutional neural networks (CNNs) have shown outstanding performance in several computer vision tasks, including image classification~\cite{he2016deep}, object detection~\cite{liu2020deep} and semantic segmentation~\cite{lateef2019survey}, many researchers have proposed to use CNNs for VPR. For instance,  S{\"u}nderhauf~\etal~\cite{sunderhauf2015performance} showed that 
features extracted from intermediate layers of CNNs trained for image classification can perform better than hand-crafted features. As a result, many have proposed to train CNNs directly for the task of place recognition~\cite{arandjelovic2016netvlad, seymour2019semantically, kim2017learned, liu2019stochastic, ge2020self}, by designing end-to-end trainable layers that can be plugged into pre-trained networks (backbones) to aggregate their rich feature maps into robust representations. These approaches demonstrated great success at large scale benchmarks \cite{torii2013visual, warburg2020mapillary} thanks to the availability of pre-trained networks and the VPR-specific datasets for fine-tuning.

Despite all the progress in the field of visual place recognition, most existing state-of-the-art  techniques either use NetVLAD~\cite{arandjelovic2016netvlad, warburg2020mapillary, hausler2021patch, yu2019spatial} or provide a variant that incorporates attention~\cite{zhang2021vector}, context~\cite{kim2017learned}, semantics~\cite{peng2021semantic} or multi-scale~\cite{hausler2021patch}. These techniques emphasize on the aggregation of local features which have proved to be invariant to viewpoint changes. However, local features are notoriously known to fail under severe illumination and seasonal changes~\cite{masone2021survey}.

Alternative approaches to NetVLAD focus on regions of interests instead of local features, by spatially pooling from the feature maps of the backbone. Such techniques include MAC (i.e., max pooling), R-MAC~\cite{tolias2015particular} and Generalized Mean (GeM)~\cite{radenovic2018fine}. Despite their performance in image retrieval~\cite{chen2021deep} these methods have been repeatedly shown to underperform NetVLAD in the task of VPR. Most recently, Berton~\etal~\cite{berton2022rethinking} proposed CosPlace, which is a variant that builds on GeM aggregator, showing strong performance on multiple VPR benchmarks.

Currently, all existing state-of-the-art techniques propose shallow aggregation layers that are plugged into very deep pre-trained backbones cropped at the last feature-rich layer. By contrast, Wang~\etal~\cite{wang2022transvpr} proposed TransVPR, a place recognition architecture that builds on the success of vision Transformers~\cite{dosovitskiy2020image} and fuse multi-level attentions to generate global and local descriptors. TransVPR achieved strong results for local feature matching. However, its global representation performance did not surpass that of NetVLAD or CosPlace. With recent advances in isotropic architectures, it has been shown that self-attention is not critical to Vision Transformers~\cite{liu2021pay}. For instance, Tolstikhin~\etal~\cite{tolstikhin2021mlp} introduced MLP-Mixer, an architecture based exclusively on multi-level perceptrons, achieving competitive results on multiple vision tasks. 

In this paper, we present MixVPR, a new holistic aggregation technique that uses feature maps extracted from a pre-entrained backbone, and iteratively incorporates global relationships into each individual feature map. It does this through a stack of isotropic blocks that we call Feature-Mixer, which consists solely of multi-layer perceptrons (MLPs).
The effectiveness of MixVPR is demonstrated by several qualitative and quantitative results where it achieves a new state-of-the-art performance on multiple benchmarks, surpassing existing techniques by a wide margin all while being extremely lightweight.


\section{Related Works}
\begin{figure*}
\begin{center}
 \includegraphics[width=0.9\linewidth]{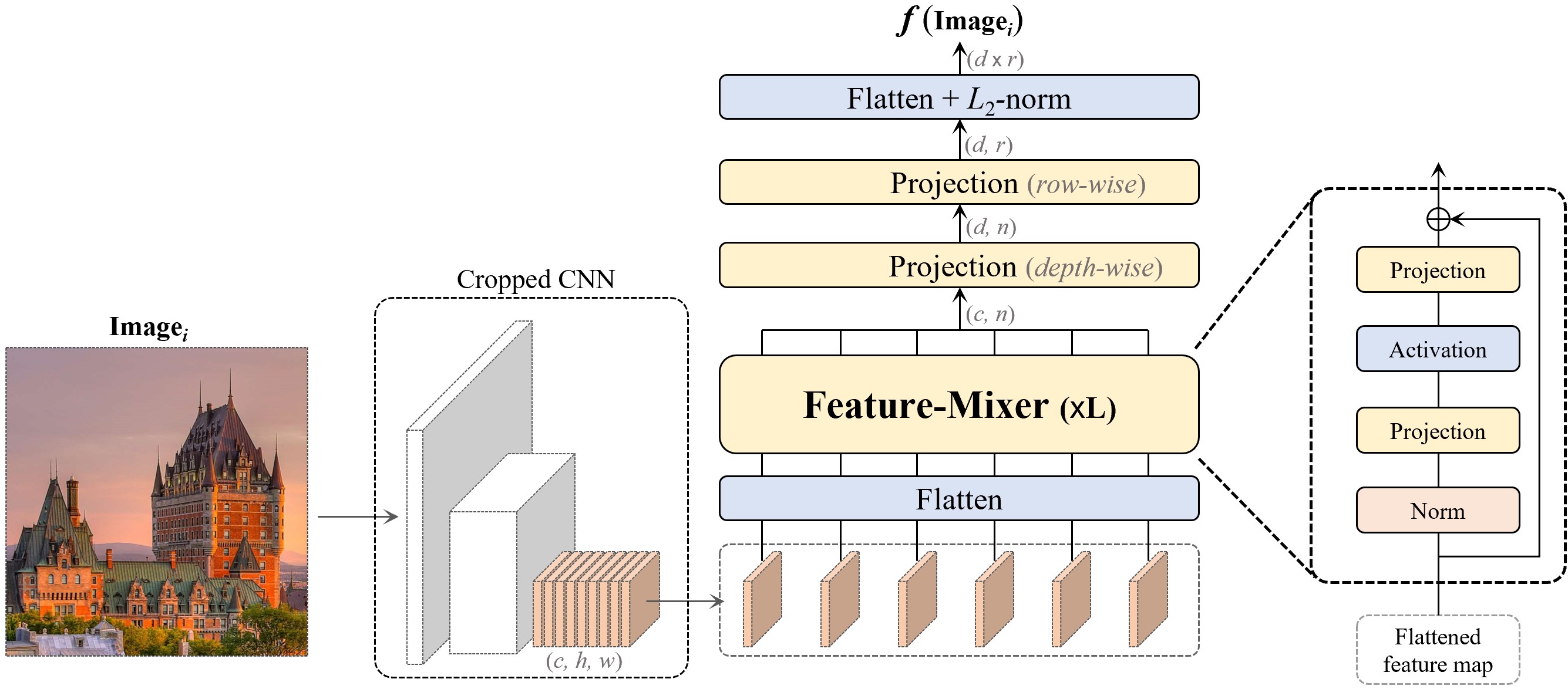}
\end{center} 
\caption{Overview of our newly proposed architecture for place recognition. MixVPR takes as input flattened feature maps from intermediate layers of a pretrained backbone. It incorporates spatial relationship in each individual feature map through a succession of Feature-Mixer blocks. The resulting output is then projected into a compact representation space and used as global descriptor.}
\label{fig:arch}
\end{figure*}

The task of visual place recognition has long been approached as an image retrieval problem, where the location of a query image is determined according to the geotags of the most relevant images retrieved from a reference database. With the success of deep learning, almost all recent VPR techniques make use of learned representations. This usually involves using features extracted from a backbone network pretrained on image classification datasets~\cite{krizhevsky2012imagenet}, followed by a trainable aggregation layer that transforms these features into robust compact representations. One notable aggregation technique is NetVLAD~\cite{arandjelovic2016netvlad}, which is a trainable variant of the VLAD descriptor, where local features are softly assigned to a learned set of clusters. As a result of the success of NetVLAD, many variants have been proposed in literature. Kim~\etal~\cite{kim2017learned} introduced Contextual Reweighting Network (CRN) which estimates a weight for each local feature from the backbone before feeding it into a NetVLAD layer; their approach introduced a slight  but consistent performance boost. Further on, SPE-VLAD~\cite{yu2019spatial} has been proposed, to enhance NetVLAD with spatial and regional features, by incorporating pyramid structure. More recently, Zhan~\etal~\cite{zhang2021vector} proposed Gated~NetVLAD, which uses a gating mechanism that incorporates attention in the computation of NetVLAD residuals.

Other techniques focus on regions of interest in the feature maps. Among the first techniques is MAC~\cite{babenko2015aggregating}, a simple aggregation method that applies max-pooling on each individual feature map, selecting only the most activated neurons. Building on that, Tolias~\etal~\cite{tolias2015particular} introduced R-MAC (Regional Maximum Activations of Convolutions) that consists of extracting multiple Region of Interest (RoI) directly from the CNN feature maps to form representations. These techniques showed impressive performance on the task of image retrieval and have since been used in VPR. Another notable aggregation technique is the Generalized Mean (GeM)~\cite{radenovic2018fine} which is a learnable generalized form of global pooling. Building on GeM, Berton~\etal~\cite{berton2022rethinking} recently proposed CosPlace, a lightweight aggregation technique that combines GeM with a linear projection layer. Their method showed impressive performance on the task of VPR, outperforming GeM and NetVLAD and achieving state-of-the-art results on multiple benchmarks.

Another trend in recent VPR works~\cite{hausler2021patch, wang2022transvpr} is to consider using a two-stage retrieval strategy, which consists of running a first global retrieval step to retrieve, for each query, the top $k$ candidates from the reference database. This step is generally more efficient because it uses $k$-NN on the global descriptors. Then, a second computationally heavy step is performed where the candidates are re-ranked according to their local features~\cite{taira2018inloc, sarlin2019coarse, sarlin2020superglue}. For instance, Patch-NetVLAD~\cite{hausler2021patch} uses NetVLAD descriptor for global description, then in a later stage, uses the local features composing NetVLAD in order to refine the retrieved candidates. This approach demonstrated good performance when re-ranking is used. Recently, TransVPR~\cite{wang2022transvpr} used a combination of CNN and Transformer by using multi-head self-attention (Transformer encoder) on top of a shallow CNN backbone. Their aim is to incorporate attention in the resulting tokens of the Transformer network. While their local feature demonstrated great performance for re-ranking, the global descriptors generated by the transformer network were not as powerful as NetVLAD or CosPlace.

In this paper, we follow recent advances in isotropic all-MLP architectures such as MLP-Mixer~\cite{tolstikhin2021mlp} and gMLP~\cite{liu2021pay}, and propose MixVPR, a novel all-MLP aggregation technique, which in contrast to TransVPR~\cite{wang2022transvpr} and Patch-NetVLAD~\cite{hausler2021patch}, does not incorporate self-attention or regional feature pooling. Although our method, MixVPR, generates global descriptors and does not perform re-ranking, it outperofms two-stage techniques such as TransVPR~\cite{wang2022transvpr}, Patch-NetVLAD~\cite{hausler2021patch} and SuperGlue~\cite{sarlin2020superglue}, while being at least $500\times$ faster in terms of latency.


\section{Methodology}
\label{sec:metho}
Our aim is to learn global compact representations that integrate features in a holistic way. Given an image $\mathcal{I}$, we first extract its feature maps $\mathbf{F} \in \mathbb{R}^{c\times h\times w}$ from the intermediate layers of a CNN backbone, $\mathbf{F} = \text{CNN}(\mathcal{I})$. Existing techniques, such as TransVPR~\cite{wang2022transvpr}, Patch-NetVLAD~\cite{hausler2021patch}, NetVLAD~\cite{arandjelovic2016netvlad}, consider $\mathbf{F}$ as a set of $c$-dimensional spatial descriptors, where each descriptor corresponds to a receptive field in the input image. 
In contrast, we consider the $3$D tensor $\mathbf{F}$ as a set of $2$D features of size $h {\times} w$ such as:
\begin{equation}
    \mathbf{F} = \{X^i\}, \quad i = \{1, \dots ,c\}
\end{equation}
 where ${X^i}$ corresponds to the $i^{th}$ activation map in $\mathbf{F}$ and sweeps across all the image (each feature map carries a certain amount of information regarding the whole image). We reshape each $2$D feature $X^i$ into a $1$D representation (flattening), resulting in flattened feature maps $\mathbf{F} \in \mathbb{R}^{c \times n}$, where $n=h {\times} w$.

Then, we feed them to what we call \emph{Feature-Mixer}, a cascade of $L$ MLP blocks of identical structure, as illustrated in Fig.~\ref{fig:arch}. Feature-Mixer takes as input a set of flattened feature maps, and incorporates global relationships into each $X^i \in \mathbf{F}$ as follows (omitting Normalization layer):
\begin{equation}
    X^i \leftarrow \mathbf{W}_2  (\sigma ( \mathbf{W}_1 \,  X^i  ) ) + X^i, \quad i=\{1, \dots, c\}
\end{equation}

where $\mathbf{W}_1$ and $\mathbf{W}_2$ are the weights of two fully-connected layers that compose the MLP, and $\sigma$ is a nonlinearity (ReLU in our case). The input to the MLP is added back to the resulting projection in a skip connection. This is proven to help the flow of gradients and further improve performance~\cite{he2016deep}.

The intuition behind Feature-Mixer is that, instead of focusing on local features, and forcing the network  to go through attention mechanism, we take advantage of the capacity of fully connected layers to automatically aggregate features in a holistic way. Feature-Mixer replaces hierarchical (pyramidal) aggregation thanks to its full receptive field, where each neuron has a glimpse into the entire input image. We use a cascade of Feature-Mixer blocks as shown in Fig.~\ref{fig:arch} in order to iteratively incorporate relationships between spatial features in each individual feature map.

For a given input $\mathbf{F} \in \mathbb{R}^{c \times n}$, Feature Mixer (FM) generates an output $\mathbf{Z} \in \mathbb{R}^{c \times n}$ of the same shape (due to its isotropic architecture), which we feed into a second Feature-Mixer block, and so on until we reach $L$ consecutive blocks, as follows:
\begin{equation}
    \mathbf{Z} = FM_L(FM_{L-1}(\dots FM_1(\mathbf{F})))
\end{equation}

$\mathbf{Z}$ is usually highly dimensional (as it has the same dimensionality as the extracted feature maps $\mathbf{F}$). To further reduce its dimensionality, we follow it by two fully connected layers that reduce its dimension depth-wise (channel-wise) then row-wise,  successively. This can be seen as a weighted pooling operation that enables control of the size of the final global descriptor.
First, we apply a depth-wise projection that maps $\mathbf{Z}$ from $\mathbb{R}^{c \times n}$ to $\mathbb{R}^{d \times n}$ as follows:
\begin{equation}
    \mathbf{Z'} = \mathbf{W}_d(Transpose(\mathbf{Z}))
\end{equation}
where $\mathbf{W}_d$ are the weights of a fully-connected layer.
We then apply a row-wise projection that maps the output $\mathbf{Z'}$ from $\mathbb{R}^{d \times n}$ to $\mathbb{R}^{d \times r}$ such as: 
\begin{equation}
    \mathbf{O} = \mathbf{W}_r(Transpose(\mathbf{Z'}))
\end{equation}
where $\mathbf{W}_r$ are the weights of another fully-connected layer. The final output $\mathbf{O}$ has a dimensionality of $d {\times} r$, which is flattened and $L_2$-normalized as usually done in VPR~\cite{arandjelovic2016netvlad, ge2020self, berton2022rethinking}.

\vspace{5pt}
\noindent\textbf{Connection to existing architectures.} Our technique is related to MLP-Mixer~\cite{tolstikhin2021mlp} where a token mixing operation is applied on spatial non-overlapping image patches. We, on the other hand, use features from CNNs that incorporate inductive bias and regard the resulting activation maps as \textit{global features}. Also, MLP-Mixer performs channel-mixing that is shared across individual spatial descriptors, which we do not employ. 

Overall, MixVPR computations are mostly matrix multiplications (of fully-connected layers) which are efficient in terms of computation compared to self-attention where the complexity scales quadratically~\cite{tolstikhin2021mlp}. Also, in MixVPR we extract feature maps from the intermediate layers (instead of the last layer) of the backbone, which reduces the number of parameters by more than half, as most parameters of a pre-trained backbone are present in the last layers.


\section{Experiments}
\begin{table*}[tbh]
\centering
\resizebox{\linewidth}{!}{%
\begin{tabular}{|l|c|ccc|ccc|ccc|ccc|}
\hline
\multirow{2}{*}{Method} & \multirow{2}{*}{dim}  & \multicolumn{3}{c|}{Pitts250k-test}           & \multicolumn{3}{c|}{MSLS-val}                 & \multicolumn{3}{c|}{SPED}                     & \multicolumn{3}{c|}{Nordland}                 \\ \cline{3-14} 
                                                &       & R@1               & R@5           & R@10            & R@1             & R@5             & R@10            & R@1             & R@5             & R@10            & R@1             & R@5             & R@10          \\ \hline \hline
AVG \cite{arandjelovic2016netvlad} $\dagger$    &$2048$ & $62.6$          & $82.7$          & $88.4$          & $59.3$          & $71.9$          & $75.5$          & $54.7$          & $72.5$          & $77.1$          & $4.4$           & $8.4$           & $10.4$          \\
GeM \cite{radenovic2018fine} $\dagger$          &$2048$ & $72.3$          & $87.2$          & $91.4$          & $65.1$          & $76.8$          & $81.4$          & $55.0$          & $70.2$          & $76.1$          & $7.4$           & $13.5$          & $16.6$          \\
NetVLAD \cite{arandjelovic2016netvlad} $\dagger$&$32768$ & $86.0$          & $93.2$          & $95.1$          & $59.5$          & $70.4$          & $74.7$          & $71.0$          & $87.1$          & $90.4$          & $4.1$           & $6.6$           & $8.2$           \\
\hline \hline
AVG~\cite{arandjelovic2016netvlad}              &$2048$ & $78.3$          & $89.8$          & $92.6$          & $73.5$          & $83.9$          & $85.8$          & $58.8$          & $77.3$          & $82.7$          & $15.3$          & $27.4$          & $33.9$          \\
GeM \cite{radenovic2018fine}                    &$2048$ & $82.9$          & $92.1$          & $94.3$          & $76.5$          & $85.7$          & $88.2$          & $64.6$          & $79.4$          & $83.5$          & $20.8$          & $33.3$          & $40.0$          \\
NetVLAD \cite{arandjelovic2016netvlad}          &$32768$ & $90.5$          & $96.2$          & $97.4$          & $82.6$          & $89.6$          & $92.0$          & $78.7$          & $88.3$          & $91.4$          & $32.6$          & $47.1$          & $53.3$          \\
SPE-NetVLAD \cite{yu2019spatial}                &$163840$ & $89.2$          & $95.3$          & $97.0$          & $78.2$          & $86.8$          & $88.8$          & $73.1$          & $85.5$          & $88.7$          & $25.5$          & $40.1$          & $46.1$          \\
Gated NetVLAD \cite{zhang2021vector}            &$32768$ & $89.7$          & $95.9$          & $97.1$          & $82.0$          & $88.9$          & $91.4$          & $75.6$          & $87.1$          & $90.8$          & $34.4$          & $50.4$          & $57.7$          \\
CosPlace \cite{berton2022rethinking}            &$2048$ & $91.5$          & $96.9$          & $97.9$          & $84.5$          & $90.1$          & $91.8$          & $75.3$          & $85.9$          & $88.6$          & $34.4$          & $49.9$          & $56.5$          \\ \hline
\textbf{MixVPR (Ours) }                         &$2048$ & $94.1$          & $98.2$          & $98.8$          & $87.0$          & $92.7$          & $94.2$          & $84.7$          & $92.1$          & $94.4$          & $57.9$          & $73.8$          & $79.0$ \\
\textbf{MixVPR (Ours) }                         &$4096$ & $\mathbf{94.6}$ & $\mathbf{98.3}$ & $\mathbf{99.0}$ & $\mathbf{88.0}$ & $\mathbf{92.7}$ & $\mathbf{94.6}$ & $\mathbf{85.2}$ & $\mathbf{92.1}$ & $\mathbf{94.6}$ & $\mathbf{58.4}$ & $\mathbf{74.6}$ & $\mathbf{80.0}$ \\ \hline
\end{tabular}%
}
\caption{\textbf{Comparison of different techniques on popular benchmarks.} $\dagger$ are results reported by the authors and confirmed using their trained networks. We however, train all six techniques on the same dataset using the same backbone network (ResNet-50). NetVLAD and its variants obtain third best performance just after the recent CosPlace method. Our technique, MixVPR, obtains by far the best performance on all benchmarks, and with big margins.}
\label{tab:sota}
\end{table*}
In this section, we run extensive experiments to show the effectiveness of the proposed MixVPR compared to existing state-of-the-art techniques by evaluating on multiple challenging benchmarks. In what follows, we present implementation details, datasets, evaluation metrics, performance comparisons and ablation studies.

\subsection{Implementation details}
\noindent\textbf{Architecture.} We implement MixVPR in PyTorch framework~\cite{paszke2019pytorch} and use existing implementations of GeM~\cite{radenovic2018fine}, NetVLAD~\cite{arandjelovic2016netvlad} and CosPlace~\cite{berton2022rethinking}. However, for techniques without existing implementation, such as SPE-NetVLAD~\cite{yu2019spatial} and Gated NetVLAD~\cite{zhang2021vector}, we do our best to faithfully reimplement them following their respective papers.
For all techniques, the CNN backbone is cropped at the last convolutional layer as recommended by their authors. MixVPR uses a backbone cropped in the middle (i.e., at the second last ResNet residual block) so that the Feature Mixer receives feature maps with a spatial dimension of $20\times 20$.
For maximum fairness, we use the exact same CNN backbone for all compared techniques (i.e., ResNet-50~\cite{he2016deep}).
The projection operation in Feature-Mixer is the Linear layer of PyTorch which we follow by a relu nonlinearity. As for the normalization layer we use LayerNorm. Finally, the output of the Feature-Mixer is projected into a smaller representation space using two consecutive fully-connected layer as described in~\ref{sec:metho}, which makes MixVPR an all-MLP architecture. Unless otherwise stated, we fix $L=4$ the number of stacked Feature-Mixer blocks.

\noindent\textbf{Training.} Using a ResNet~\cite{he2016deep} backbone pre-trained on ImageNet~\cite{krizhevsky2012imagenet}, we train all techniques on the same dataset, following the standard framework of GSV-Cities~\cite{gsvcities}, which proposes a highly accurate dataset of $67$k places depicted by $560$k images. For the loss function, we use Multi-Similarity loss~\cite{wang2019multi} as it has been shown to perform best for visual place recognition~\cite{gsvcities}.
We use batches containing $P=120$ places, each depicted by $4$ images resulting in mini-batches of $480$ images. We use Stochastic Gradient Descent (SGD) for optimization, with momentum $0.9$ and weight decay of $0.001$. The initial learning rate of $0.05$ is divided by $3$ after each $5$ epochs. Finally, we train for a maximum of $30$ epochs using images resized to $320 {\times} 320$.

\noindent\textbf{Evaluation.} For evaluation we use the following $5$ benchmarks. Pitts250k-test~\cite{torii2013visual}, which contains $8$k queries and $83$k reference images, collected from Google Street View and Pitts30k-test~\cite{torii2013visual} which is a subset of Pitts250k and comprises $8$k queries and $8$k references. Both Pittsburgh datasets show significant viewpoint changes. SPED~\cite{zaffar2021vpr} benchmark contains $607$ queries and $607$ references from surveillance cameras presenting significant seasonal and illumination variations. MSLS~\cite{warburg2020mapillary} benchmark has been collected using car dashcams and presents a wide range of viewpoint and illumination changes. Finally, Nordland~\cite{zaffar2021vpr} is an extremely challenging benchmark which has been collected in $4$ seasons using a camera mounted in front of a train, it comprises scenes ranging from snowy winter to sunny summer with extreme appearance changes.
We follow the same evaluation metric of \cite{arandjelovic2016netvlad, kim2017learned, warburg2020mapillary, zaffar2021vpr, wang2022transvpr, berton2022rethinking}, where the recall@k is measured. The query image is determined to be successfully retrieved if at least one of the top-$k$ retrieved reference images is located within $d = 25$ meters from the query one.

\subsection{Comparison to the state of the art}

In this section, we compare the performance of MixVPR against existing methods in visual place recognition on $4$ challenging benchmarks. We compare against AVG~\cite{arandjelovic2016netvlad}, GeM~\cite{radenovic2018fine}, NetVLAD~\cite{arandjelovic2016netvlad} and two of its recent variants SPE-VLAD~\cite{yu2019spatial} and Gated NetVLAD~\cite{zhang2021vector}, and CosPlace which recently demonstrated state-of-the-art performance. 
Results are shown  in Table~\ref{tab:sota}. The lines with the sign $\dagger$ are performance of AVG, GeM and NetVLAD  trained on Pitts30k-train dataset. For fair comparison, we re-train them using the same backbone and dataset as our technique. Results are shown in the rest of the table. As can be seen, our method convincingly outperforms all other techniques on all benchmarks with a large margin. For instance, MixVPR achieves a new all-time high recall@1 of $\mathbf{94.6}$\% on Pitts250k-test which is $3.1$ percentage points increase over the recent CosPlace technique and over $4.1$ points increase compared to NetVLAD.

On MSLS, performance is even more interesting, where we achieve $\mathbf{88.0}$\% recall@1, which, to the best of our knowledge, is the best score ever achieved. This is $3.5$ and $5.4$ percentage points increase over CosPlace and NetVLAD which achieved $84.5$\% and $82.6$\% recall@1 respectively. This showcases the effectiveness of our technique on datasets presenting a lot of viewpoint variations.

On SPED benchmark, where places exhibit  drastic appearance change due to seasonal changes and day-night illumination, our technique surpasses all other techniques achieving $\mathbf{85.2}$\% recall@1, which is $7.5$ points more than NetVLAD, the second best performing technique on SPED. 

Finally and most importantly, on the extremely challenging Nordland benchmark, MixVPR achieves $69$\% and and $79$\% relative improvement over CosPlace and NetVLAD ($\mathbf{58.4}$\% vs $34.4$\% and $32.6$\% resp.), and more than double compared to the rest of the other techniques.

\subsection{Comparing against two-stage techniques} 
\begin{table}[t]
\centering
\resizebox{\columnwidth}{!}{%
\begin{tabular}{|l|c|c|ccc|}
\hline
\multirow{2}{*}{Method}               & \multirow{2}{*}{\begin{tabular}[c]{@{}c@{}}Extraction \\ latency (ms)\end{tabular}} & \multirow{2}{*}{\begin{tabular}[c]{@{}c@{}}Matching\\ latency (s)\end{tabular}} & \multicolumn{3}{c|}{Mapillary Challenge}       \\ \cline{4-6} 
                                      &                                                                                     &                                                                                 & \footnotesize{R@1}      & \footnotesize{R@5}           & \footnotesize{R@10}         \\ \hline \hline
Super-Glue~\cite{sarlin2020superglue} & $160$                                                                                 & $7.5 $                                                                            & $50.6$          & $56.9$          & $58.3 $         \\
DELG~\cite{cao2020unifying}           & $190 $                                                                                & $35.2 $                                                                           & $52.2$          & $61.9$          & $65.4 $         \\
Patch-NetVLAD~\cite{hausler2021patch} & $1300 $                                                                               & $7.4 $                                                                            & $48.1$          & $59.4 $         & $62.3 $         \\
TransVPR~\cite{wang2022transvpr}      & $45  $                                                                                & $3.2 $                                                                            & $63.9 $         & $74.0 $         & $77.5 $         \\ \hline
NetVLAD~\cite{arandjelovic2016netvlad}& $17  $                                                                                 & $-$                                                                              & $35.1$ & $47.4$ & $51.7$ \\
\textbf{MixVPR (Ours)}                & $6  $                                                                                 & $- $                                                                              & $\mathbf{64.0}$ & $\mathbf{75.9}$ & $\mathbf{80.6}$ \\ \hline
\end{tabular}%
}
\caption{\textbf{Comparison with two-stage retrieval techniques.} The first four techniques use a second refinement pass (matching) to re-rank the top candidates in order to improve retrieval performance. MixVPR (ours) does not use re-ranking, which makes it at least $500\times$ faster all while outperforming existing state-of-the-art. (a NVIDIA Titan Xp has been used to calculate latency).}
\label{tab:two-stage}
\end{table}
Some techniques use a two-stage retrieval framework, where a first pass is performed to retrieve the best $M$ candidates using global representations, then a second pass (re-ranking) is executed to perform geometric verification on the local features between the query and each one of the $M$ candidates~\cite{wang2022transvpr}. This is known to increase recall@N performance at the expense of heavy computation and memory overhead. We compare against Patch-NetVLAD~\cite{hausler2021patch}, DELG~\cite{cao2020unifying}, SuperGlue~\cite{sarlin2020superglue} and TransVPR~\cite{hausler2021patch} which are state-of-the-art techniques that perform two-stage visual place recognition. Table~\ref{tab:two-stage} shows performance on the Mapillary Challenge. Although our technique does not perform any re-ranking, it achieves better performance than existing two-stage techniques while being orders of magnitudes more efficient in terms of memory and computation (over $500\times$ faster retrieval time). We believe that MixVPR can replace two-stage techniques in applications where time and resources are of great importance. For instance, MixVPR takes only $6$ milliseconds to generate an image representation, while the second fastest method, TransVPR, takes 45 milliseconds. Matching latency does not apply to MixVPR since it is a global technique that does not perform re-ranking. However, it is clear from Table~\ref{tab:two-stage} that the re-ranking phase takes a lot of time, making such techniques unusable in real-time applications.

\subsection{Ablation studies}\label{sec:exp:ablation}
We conduct multiple ablation experiments to further validate the design of MixVPR.

\subsubsection{Hyperparameters}
\begin{table}[t]
\centering
\resizebox{\columnwidth}{!}{%
\begin{tabular}{|c|c|c|ccc|ccc|}
\hline
\multirow{2}{*}{\begin{tabular}[c]{@{}c@{}}$\times L$\end{tabular}} & \multirow{2}{*}{\begin{tabular}[c]{@{}c@{}}\# params\\ (M)\end{tabular}} & \multirow{2}{*}{\begin{tabular}[c]{@{}c@{}}Latency\\ (ms)\end{tabular}} & \multicolumn{3}{c|}{Pitts30k-test} & \multicolumn{3}{c|}{MSLS-val} \\ \cline{4-9} 
                                                                    &                                                                                &                                                                         & \footnotesize{R@1}           & \footnotesize{R@5}           & \footnotesize{R@10}      & \footnotesize{R@1}           & \footnotesize{R@5}           & \footnotesize{R@10}    \\ \hline \hline
$0$                                                                   & $9.6$                                                                          & $6.3$                                                                    & $89.5 $      & $95.0 $     & $96.2 $     & $82.9 $    & $90.7 $    & $91.9 $   \\
$1$                                                                   & $9.9$                                                                          & $6.5 $                                                                    & $91.3 $      & $95.6 $     & $96.5 $     & $86.9 $    & $92.8 $    & $94.3 $   \\
$2$                                                                   & $10.2 $                                                                          & $6.6 $                                                                    & $91.3 $      & $95.8 $     & $96.6 $     & $87.6 $    & $93.1 $    & $94.6 $   \\
$4$                                                                   & $10.9 $                                                                          & $6.6 $                                                                    & $91.9 $      & $\mathbf{95.9}$     & $\mathbf{96.7} $     & $\mathbf{87.6} $    & $\mathbf{93.5} $    & $\mathbf{95.0} $   \\
$8$                                                                   & $12.2 $                                                                          & $7.2 $                                                                    & $\mathbf{92.3}$      & $95.9 $     & $96.6 $     & $87.2 $    & $92.6 $    & $93.9 $   \\ \hline
\end{tabular}%
}
\caption{\textbf{Ablation on the number of Feature-Mixer blocks.} The baseline ($L=0$) does not use Feature-Mixer. We compare it to different configurations by varying $L$ the number of stacked Feature-Mixer blocks. Overall, $L=4$ stacks of Feature-Mixer performs the best on all benchmarks.}
\label{tab:feature-mixer}
\end{table}
In order to showcase the effect of Feature-Mixer, we conduct multiple experiments by varying $L$ the number of Feature-Mixer blocks. First, we train a baseline network  without Feature-Mixer ($L=0$), and compare its performance when trained with multiple stacked Feature-Mixer blocks ($L \in \{1,2,4,8\}$). Results are shown in Table~\ref{tab:feature-mixer}, where we see that introducing only one Feature-Mixer layer improves recall@1 performance by $1.8$ recall@1 points from $89.5$\% to $91.3$\% on Pitts30k-test and $4$ on MSLS  from $82.9$\% to $86.9$\%. Overall, the best results are obtained with $4$ Feature-Mixer layers, although all configurations achieve similar performance. Feature-Mixer adds $340$k parameters to the network, therefore we can refer to Table~\ref{tab:feature-mixer} to choose the best compromise.

\subsubsection{Descriptor dimensionality}
\begin{figure}[thb]
\begin{center}
\includegraphics[width=0.9\linewidth]{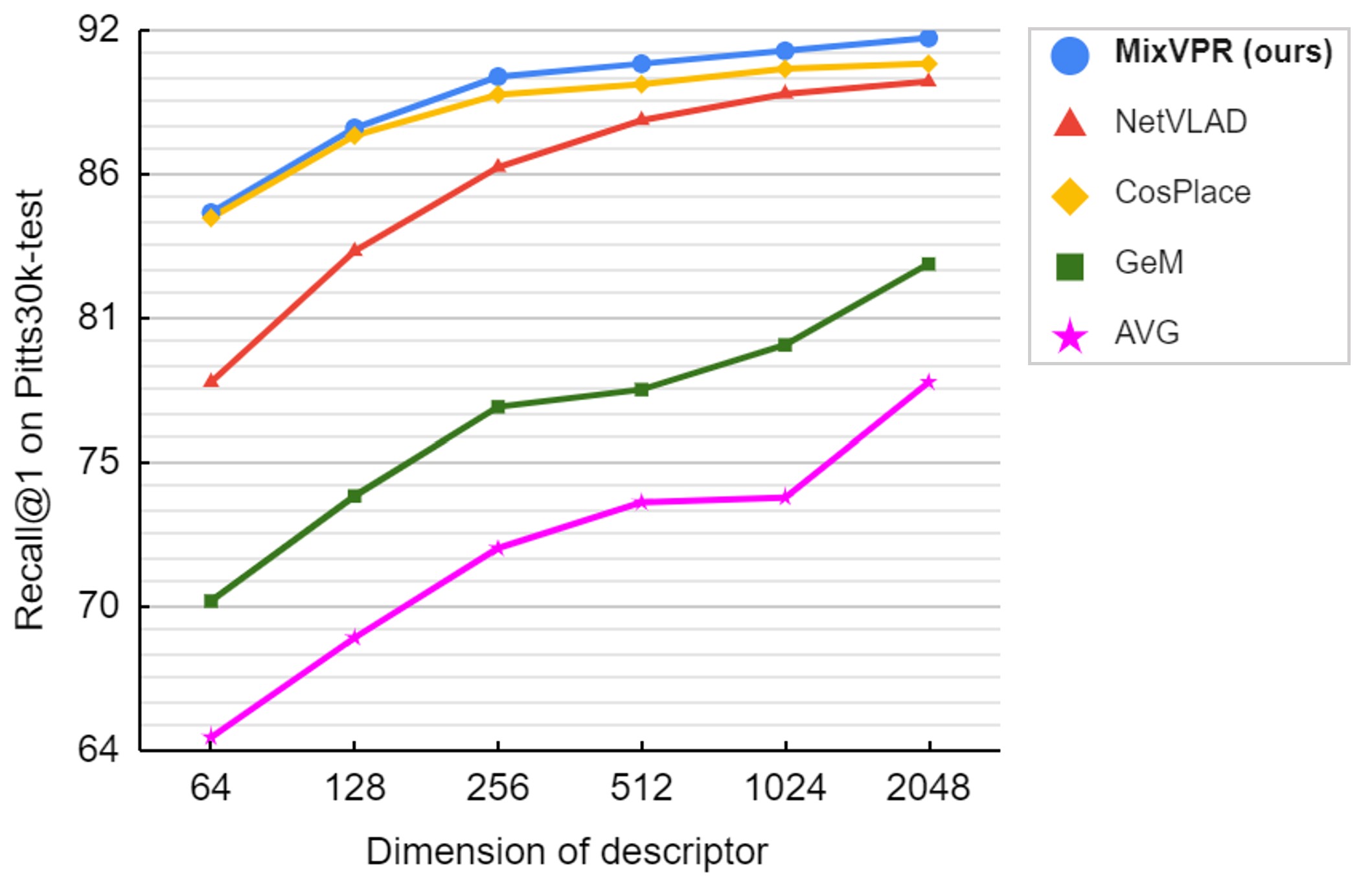}
\end{center}
   \caption{Recall@1 performance on Pitts30k-test with different dimensionality configurations.}
\label{fig:dim}
\end{figure}
The architecture of MixVPR allows to configure the dimensionality of the output descriptor, by fixing the size of the last two projection operations. In ~\ref{fig:dim} we show recall@1 performance for different dimensionality configurations on Pitts30k-test. For NetVLAD, GeM and AVG, we used PCA dimensionality reduction learned on a subset of $10$k images from the training set. CosPlace, like MixVPR, allows to configure the output dimensionality. Hence, we trained once for each configuration. From the chart in Fig.~\ref{fig:dim}, we can clearly see that MixVPR outperforms all other techniques.

\subsubsection{Backbone architecture}
\begin{table}[t]
\centering
\resizebox{\columnwidth}{!}{%
\begin{tabular}{|l|c|ccc|ccc|}
\hline
\multirow{2}{*}{Backbone} & \multirow{2}{*}{\begin{tabular}[c]{@{}c@{}}\# param. \\ (M)\end{tabular}} & \multicolumn{3}{c|}{Pitts30k-test} & \multicolumn{3}{c|}{MSLS-val} \\ \cline{3-8} 
                          &                                                                               & \footnotesize{R@1}           & \footnotesize{R@5}           & \footnotesize{R@10}      & \footnotesize{R@1}           & \footnotesize{R@5}           & \footnotesize{R@10}    \\ \hline \hline
ResNet-18                 & $3.5$                                                                          & $89.5$       & $95.0$      & $96.2$      & $82.7$     & $89.1$     & $91.8$    \\
ResNet-34                 & $8.2$                                                                          & $90.5 $      & $95.2 $     & $96.3$      & $85.3$     & $91.6$     & $93.4$    \\
ResNet-50                 & $10.9$                                                                          & $91.6$      & $96.0 $     & $96.7$      & $88.0$     & $92.8$     & $94.5$    \\
ResNeXt-50                & $10.9$                                                                          & $91.7$     & $95.7 $     & $96.5$      & $87.0$     & $93.5$     & $94.7 $   \\ \hline
\end{tabular}%
}
\caption{\textbf{Comparing different backbones.} Each backbone is cropped at the fourth residual block (before the last one), which results in half the number of parameters of the same backbone used in CosPlace or netVLAD. MixVPR only needs intermediate features of the backbone.}
\label{tab:backbone}
\end{table}
In Table~\ref{tab:backbone} we conduct multiple experiments using different backbone architectures. Since we crop the backbone at the $4^{th}$ residual layer (instead of the last) we end up cropping out half the total number of parameters, thus accelerating computation and reducing memory use. As can be seen in Table~\ref{tab:backbone}. Using ResNet-18~\cite{he2016deep} we end up with only $3.5$M parameters, which is 15\% the number of parameters in CosPlace or NetVLAD, all while getting competitive results. We believe ResNet-18 can be used in applications where real-time is top priority. Importantly, MixVPR obtains state-of-the-art performance using only ResNet-34 which comprises $70$\% less parameters compared to CosPlace while outperforming it by $2.3$ recall@1 points on MSLS. The best overall results are obtained with ResNet-50 where the number of parameters ($10.9$M) is less than half that of NetVLAD or CosPlace. Interestingly, using ResNeXt50~\cite{xie2017aggregated} did not increase performance compared to ResNet-50. We believe this is because MixVPR  draws much of its performance from the Feature Mixing rather than the backbone network.

\subsection{Qualitative Results}
\begin{figure*}
\begin{center}
\includegraphics[width=0.8\linewidth]{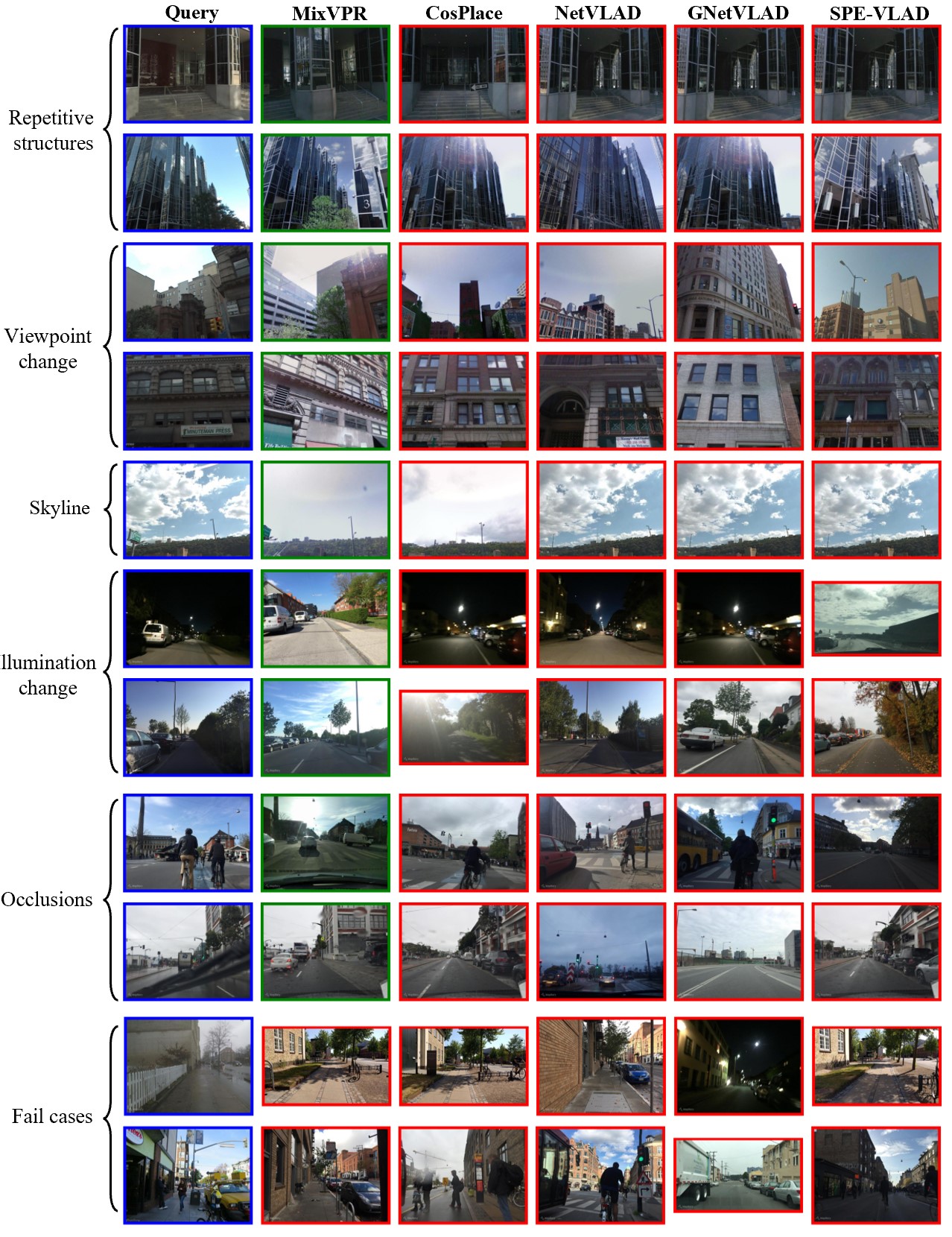}
\end{center}
   \caption{Comparison of challenging retrieval scenarios on MSLS and Pitts30k datasets. MixVPR succeeds the retrieval of all these challenging queries, while all other techniques fail. This qualitative results highlight the robustness of MixVPR to extreme scenarios.}
\label{fig:fails}
\end{figure*}

Fig.~\ref{fig:fails} illustrates qualitative results of the retrieval of some challenging queries. We discuss $5$ scenarios where all other techniques struggle retrieving the correct match while MixVPR succeeds. \textbf{Repetitive structures:} this is a serious problem for VPR techniques, since different places may contain the same type of building or structure with the same layout or texture, this can fool the recognition system and induce a lot of false positives as we can see in the first two rows of Fig.~\ref{fig:fails}, where only MixVPR succeeded in retrieving the right reference, while all other techniques retrieved images of different places that are overly similar to the query. \textbf{Viewpoint change:} for this scenario, techniques that focus on local features, such as NetVLAD, tend to perform better. However, in rows~3-4 of Fig~\ref{fig:fails}, only MixVPR retrieved the right references, which highlights its capacity to deal with extreme viewpoint changes. \textbf{Skyline:} some environments contain few static structures such as buildings and poles, making the image lack distinctive textures. In this case, the skyline constitutes an important signature of the place. As we can see in row~5 of Fig~\ref{fig:fails}, only MixVPR succeeded in retrieving the correct reference based most likely on the skyline all while ignoring the cloud texture. \textbf{Illumination change:} we believe this to be the most important aspect of a robust VPR system, because illumination variations occur on a daily basis, such an example is illustrated in rows~6-7 of Fig~\ref{fig:fails} where the query is taken during the night and its reference is taken during the day. CosPlace, NetVLAD and Gated NetVLAD all retrieved images of locations taken at nighttime, in contrast, MixVPR retrieved the correct reference even though it is visually very tricky even for the human eye. This highlights the robustness of our method in extremely challenging situations. \textbf{Occlusions:} this can be challenging when part of the image is obstructed with an object that can affect the global semantic of the image. For instance, row~8 of Fig~\ref{fig:fails} shows a query with two cyclists in the middle of the field of view (FoV), which tricked other techniques to retrieve the wrong references containing cyclists in the middle of the FoV. Only MixVPR ignored the cyclists and successfully retrieved the right reference. Finally, we show two cases where all techniques fail, due to extreme environmental changes and the presence of a lot of  occlusions.

\subsubsection{Visualizing learned weights}
\begin{figure}
\begin{center}
\includegraphics[width=0.8\linewidth]{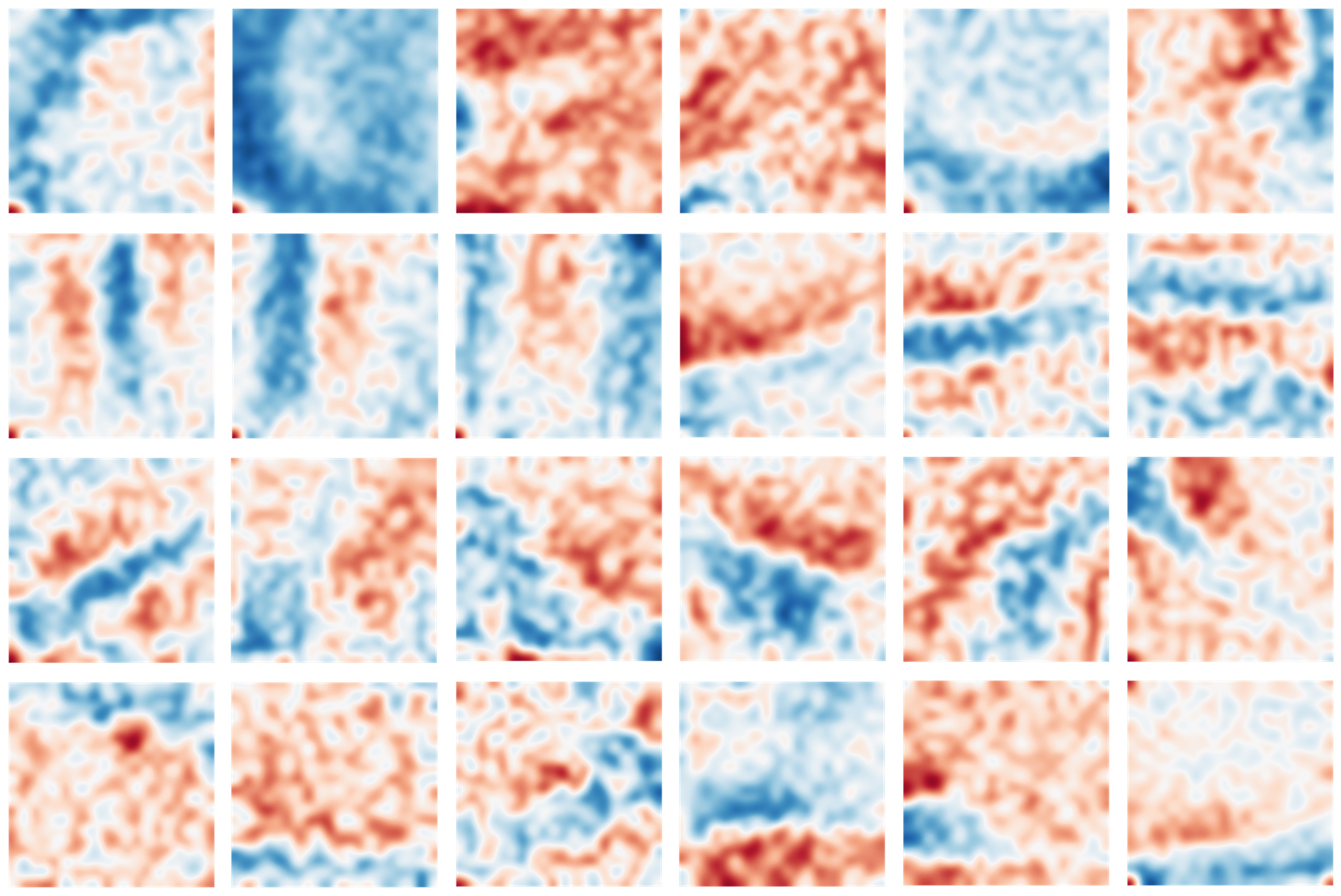}
\end{center}
   \caption{Illustration of learned weights from a subset of $24$ neurons from the first Feature-Mixer block. Blue color corresponds to positive weights and Red corresponds to negative weights.}
\label{fig:neurones}
\end{figure}
Fig~\ref{fig:neurones} illustrates a subset of learned weights from the first hidden layer of Feature-Mixer ($24$ neurons out of $400$). The weights of each unit have been reshaped to $20 \times 20$ to match the spatial size of feature maps coming from the backbone. As we can see, hidden units in Feature-Mixer learned a wide range of regional feature selection. We observe that some neurons focus on one or multiple small spots of the image, while other focus on the entire input. We believe the combination of these neurons can replace attention and pyramidal scheme in deep model for VPR.

\section{Conclusion}
In this work, we designed a novel all-MLP aggregation technique that employs feature maps from pretrained networks, and learns robust representations in a cascade of feature mixing. MixVPR is composed of a stack of Feature-Mixer blocks, where each block incorporates global relationships between individual feature maps. We demonstrated the effectiveness of the feature mixing through ablation studies, and showed that MixVPR outperforms existing state-of-the-art by a wide margin on every benchmark we tested on. Finally, we also compared performance of MixVPR against two-stage retrieval techniques such as Patch-NetVLAD and TransVPR and showed that our method is superior while being over $500\times$ faster.

\vspace{5pt}
\noindent\textbf{Acknowledgement:} This work has been supported by The Fonds de Recherche du Québec Nature et technologies (FRQNT). We gratefully acknowledge the support of NVIDIA Corporation with the donation of a Quadro RTX 8000 GPU used for our experiments.

{\small
\bibliographystyle{ieee_fullname}
\bibliography{bibliography}
}

\end{document}